\documentclass[conference,usletter]{IEEEtran}
\usepackage{times,amsmath,amssymb}
\usepackage{slashbox}
\usepackage{graphicx}
\usepackage{xcolor}
\usepackage{algorithm}
\usepackage{algorithmicx}
\usepackage{adjustbox}
\usepackage{setspace}
\usepackage{multirow}
\usepackage[noend]{algpseudocode}

\newtheorem{defn}{Definition}
\begin{document}

\title{Multilinear Class-Specific Discriminant Analysis}
\author{\IEEEauthorblockN{Dat Thanh Tran \IEEEauthorrefmark{1}, Moncef Gabbouj\IEEEauthorrefmark{1} and Alexandros Iosifidis\IEEEauthorrefmark{2}}
\IEEEauthorblockA{\IEEEauthorrefmark{1}Laboratory of Signal Processing, Tampere University of Technology, Tampere, Finland\\
\IEEEauthorrefmark{2}Department of Engineering, Electrical and Computer Engineering, Aarhus University, Denmark\\
Email: \{dat.tranthanh, moncef.gabbouj\}@tut.fi, alexandros.iosifidis@eng.au.dk}

}
\maketitle

\begin{abstract}
There has been a great effort to transfer linear discriminant techniques that operate on vector data to high-order data, generally referred to as Multilinear Discriminant Analysis (MDA) techniques. Many existing works focus on maximizing the inter-class variances to intra-class variances defined on tensor data representations. However, there has not been any attempt to employ class-specific discrimination criteria for the tensor data. In this paper, we propose a multilinear subspace learning technique suitable for applications requiring class-specific tensor models. The method maximizes the discrimination of each individual class in the feature space while retains the spatial structure of the input. We evaluate the efficiency of the proposed method on two problems, i.e. facial image analysis and stock price prediction based on limit order book data.
\end{abstract}

\section{Introduction}
Over the past two decades, several subspace learning techniques have been proposed for computer vision and pattern recognition problems. The aim of subspace learning is to find a set of bases that optimizes a given objective function (or criterion) enhancing properties of interest in the learnt subspace. The obtained projection can be subsequently used either as a means of preprocessing or as a classifier. One of the methods used as a preprocessing step is Principal Component Analysis (PCA) (\cite{wold1987principal}), which finds a projection maximizing the data dispersion. While PCA retains most spectral information, it is an unsupervised method that does not utilize labeling information to increase class discrimination. Linear Discriminant Analysis (LDA) (\cite{duda2001pattern}) is one of the most widely used discriminant learning techniques due to its success in many applications (\cite{iosifidis2012activity,iosifidis2012multi,ren2016enhanced}). In LDA, each class is assumed to follow a unimodal distribution and is represented by the corresponding class mean vector. LDA optimizes the ratio between inter-class and intra-class scatters. Several extensions have been proposed in order to relax these two assumptions (\cite{zhu2006subclass,iosifidis2013optimal,iosifidis2014kernel}). An important characteristic of LDA is that the maximal dimensionality of the learnt subspace is limited by the number of classes $C$ forming the problem at hand. For problems where the objective is to discriminate one class from all other alternatives, i.e. for binary problems like face verification, this might not be an optimal choice for class discrimination.

To tackle the latter limitation of LDA, class-specific discriminant analysis techniques were proposed (\cite{goudelis2007class,zafeiriou2012regularized,arashloo2014class,iosifidis2016class}). In the class-specific setting, a unimodal distribution is still assumed for the class of interest (hereafter noted as positive class), and the objective is to determine class-specific projections discriminating the samples forming the positive class from the rest samples (forming the negative class) in the subspace. By defining suitable out-of-class and in-class scatter matrices, the maximal subspace dimensionality is limited by the cardinality of the positive class, leading to better class discrimination and classification (\cite{goudelis2007class,arashloo2014class,iosifidis2015class}). Various extensions have been proposed to utilize class-specific formulation. For example, (\cite{iosifidis2015class}) proposed a solution to optimize both the class representation and the projection; in addition, approximate and incremental learning solutions were proposed  in (\cite{iosifidis2016class,iosifidis2016tifs}).

While being able to overcome the limitation in subspace dimensionality of LDA, there is yet a limitation in the existing Class-Specific Discriminant Analysis (CSDA) methods. These methods are defined on vector data. Since many types of data are naturally represented by (high-order) matrices, generally referred as tensors, exploiting vector-based learning approaches might lead to the loss of spatial information being available on the data. For example, a grayscale image is naturally represented as a matrix (i.e. second order tensor), a color image is represented as a third order tensor and a multi-dimensional time series is represented as a third order tensor. Vectorizing such high-order tensors results to high-dimensional vectors, leading to high computational costs and the small sample size problem (\cite{chen2000SSS}). In order to address such issues, generalizations of many linear subspace learning methods to multilinear ones have been proposed, including MPCA (\cite{lu2008mpca}) and CMDA (\cite{li2014multilinear}) as the multilinear extensions of PCA, GTDA (\cite{tao2007general}) and DATER (\cite{yan2005discriminant}) as the multilinear extensions of LDA.

With the potential advantage of using tensorial data representations in (binary) verification problems, in this work, we propose to extend the class-specific discrimination criterion for tensor-based learning and formulate the Multilinear Class-Specific Discriminant Analysis (MCSDA) method. Moreover, we provide a time complexity analysis for the proposed method and compare it with its vector counterparts. We conducted experiments in two problems involving data naturally represented in a tensor form, i.e. facial image analysis and stock price prediction based on limit order book data. Experimental results show that the proposed MCSDA is able to outperform related tensor-based and vector-based methods and to compare favourably with recent methods.

The rest of the paper is organized as follows. Section 2 introduces the notations used throughout the paper, as well as related prior works. In section 3, we formulate the proposed MCSDA method and provide our analysis on its time complexity. Section 4 presents our experimental analysis, and conclusions are drawn in Section 5.

\section{Notations and prior work}\label{S:previouswork}
We start by introducing the notations used throughout the paper and related definitions from multilinear algebra. In addition, previous works in discriminant analysis utilizing multi-class and class-specific criteria are briefly reviewed.

\subsection{Multilinear Algebra Concepts}
In this paper, we denote scalar values by either low-case or upper-case characters $(x, y, X, Y \dots)$, vectors by low-case bold-face characters $(\mathbf{x}, \mathbf{y}, \dots)$, matrices by upper-case bold-face characters $(\mathbf{A}, \mathbf{B}, \dots)$ and tensors by calligraphic capital characters $(\mathcal{X}, \mathcal{Y}, \dots)$. A tensor is a multilinear matrix with $K$ modes, and is defined as $\mathcal{X} \in \mathbb{R}^{I_1 \times I_2 \times \dots \times I_K}$, where $I_{k}$ denotes the dimension in mode-$k$. The entry in the $i_k$th index in mode-$k$ for $k=1,\dots, N$ is denoted as $\mathcal{X}_{i_1,i_2,\dots,i_K}$.

\begin{defn}[Mode-$k$ Fiber and Mode-$k$ Unfolding]\label{def1}
	The mode-$k$ fiber of a tensor $\mathcal{X} \in \mathbb{R}^{I_1 \times I_2 \times \dots \times I_K}$ is a vector of $I_k$-dimensional, given by fixing every index but $i_k$. The mode-$k$ unfolding of $\mathcal{X}$, also known as mode-$k$ matricization, transforms the tensor $\mathcal{X}$ to matrix $\mathbf{X}_{(k)}$, which is formed by arranging the mode-$k$ fibers as columns. The shape of $\mathbf{X}_{(k)}$ is $\mathbb{R}^{I_k \times I_{\bar{k}}}$ with $I_{\bar{k}}=\prod_{i=1,i \neq k}^{K} I_i$.
\end{defn}

\begin{defn}[Mode-$k$ Product]\label{def2}
	The mode-$k$ product between a tensor $\mathcal{X}=[x_{i_1},\dots , x_{i_K}] \in  \mathbb{R}^{I_1 \times \dots I_K}$ and a matrix $\mathbf{W}\in \mathbb{R}^{J_{k}\times I_k}$ is another tensor of size $I_1\times \dots \times J_{k}\times \dots \times I_K$ and denoted by $\mathcal{X} \times_{k} \mathbf{W}$. The element of $\mathcal{X} \times_{k} \mathbf{W}$ is defined as $[\mathcal{X}\times_{k}\mathbf{W}]_{i_1, \dots , i_{k-1}, j_k, i_{k+1},\dots, i_K}=\sum_{i_k=1}^{I_K}[\mathcal{X}]_{i_1,\dots,i_{k-1},i_k,\dots, i_K}[\mathbf{W}]_{j_k,i_k}$.
\end{defn}

With the definition of mode-$k$ product and mode-$k$ unfolding, the following equation holds
\begin{equation}\label{eq1}
(\mathcal{X}\times_k\mathbf{W})_{(k)} = \mathbf{W}\mathbf{X}_{(k)}
\end{equation}
For convenience, we denote $\mathcal{X}\times_1\mathbf{W}_1\times\dots\times_K
\mathbf{W}_K$ by $\mathcal{X} \prod_{k=1}^{K}\times_k\mathbf{W}_k$.

\subsection{Linear Discriminant Analysis}
Let us denote by $\mathbf{X}=[\mathbf{x}_1,\dots ,\mathbf{x}_N]\in \mathbb{R}^{D\times N}$ a set of $N$ $D$-dimensional vectors, each of which has an associated class label $l_j$ ($j=1,\dots,N$) belonging to the label set $\{c_i \mid i=1,\dots,C\}$. $n_i$ is the number of samples in class $c_i$. Let $\mathbf{x}_{i,j}$ denote the $j$th sample of class $c_i$. The mean vector of class $c_i$ is calculated as $\mathbf{m}_i=\frac{1}{n_i}\sum_{j=1}^{n_i}\mathbf{x}_{i,j}$. The mean vector of the entire set is $\mathbf{m}=\frac{1}{N}\sum_{i=1}^{C}\sum_{j=1}^{n_i}\mathbf{x}_{i,j}=\frac{1}{N}\sum_{i=1}^{C}n_{i}\mathbf{m}_{i}$. Linear Discriminant Analysis (LDA) seeks an othonormal projection matrix $\mathbf{W}\in \mathbb{R}^{D\times d}$ that maps each sample $\mathbf{x}_i$ to a lower $d$-dimensional feature space $(d<D)$ in which samples from different classes are highly discriminated. $\mathbf{W}$ is obtained by maximizing the ratio between the inter-class and intra-class variances in the feature space (\cite{welling2005fisher}), i.e.
\begin{equation}\label{eq2}
J(\mathbf{W}) =\frac{\sum_{i=1}^{C}n_i\big\Vert\mathbf{W}^{T}\mathbf{m}_i-\mathbf{W}^{T}\mathbf{m}\big\Vert_{F}^{2}}{\sum_{i=1}^{C}\sum_{j=1}^{n_i}\big\Vert \mathbf{W}^{T}\mathbf{x}_{i,j}-\mathbf{W}^{T}\mathbf{m}_i\big\Vert_{F}^{2}} = \frac{tr(\mathbf{W}^{T}\mathbf{S}_b\mathbf{W})}{tr(\mathbf{W}^{T}\mathbf{S}_w\mathbf{W})}
\end{equation}
where $\mathbf{S}_{b}=\sum_{i=1}^{C}n_i(\mathbf{m}_i-\mathbf{m})(\mathbf{m}_i-\mathbf{m})^{T}$ denotes the between-class scatter matrix and $\mathbf{S}_{w}=\sum_{i=1}^{C}\sum_{j=1}^{n_i}(\mathbf{x}_{i,j}-\mathbf{m}_{i})(\mathbf{x}_{i,j}-\mathbf{m}_{i})^{T}$ denotes the within-class scatter matrix. By maximizing $J(\mathbf{W})$ in (\ref{eq2}), the dispersion between the data and the corresponding class mean is minimized while the dispersion between each class mean and the total mean is maximized in the projected subspace. The columns of $\mathbf{W}^{*}$ are formed by the eigenvectors corresponding to the $d \leq C-1$ largest eigenvalues of $\mathbf{S}_{w}^{-1}\mathbf{S}_{b}$.

\subsection{Class-Specific Discriminant Analysis}
While LDA seeks to project all data samples to a common subspace where the data samples between classes are expected to be highly discriminated, class-specific discriminant analysis (CSDA) learns a subspace discriminating the class of interest from everything else. For a $C$-class classification problem, application of CSDA leads to the determination of $C$ different discriminant subspaces $\mathbb{R}^{d_i}, d_i<D, i=1,\dots,C$ in an One-versus-Rest manner, where $d_i$ is the dimensionality of the $i$th subspace that discriminates samples of class $c_i$ from the rest.

Let us denote $p$, $n$ the positive and negative labels, respectively. The optimal mapping $\mathbf{W}$ is obtained by maximizing the following criterion
\begin{equation}\label{eq4}
J(\mathbf{W})=\frac{D_O}{D_I}
\end{equation}
where $D_O=\sum_{j,l_j\neq p}\big\Vert \mathbf{W}^{T}\mathbf{x}_j-\mathbf{W}^{T}\mathbf{m}_p\big\Vert_{F}^{2}$ is the out-of-class distance and $D_I=\sum_{j,l_j=p}\big\Vert \mathbf{W}^{T}\mathbf{x}_j-\mathbf{W}^{T}\mathbf{m}_p\big\Vert_{F}^{2}$ is the in-class distance, respectively. That is the positive class is assumed to be unimodal and the optimal projection matrix $\mathbf{W}$ maps the positive class vectors as close as possible to the positive class mean $\mathbf{m}_p$ while keeping the negative samples far away from $\mathbf{m}_p$ in the subspace. $J(\mathbf{W})$ in (\ref{eq4}) can be expressed as
\begin{equation}\label{eq5}
J(\mathbf{W})=\frac{tr(\mathbf{W}^{T}\mathbf{S}_{O}\mathbf{W})}{tr(\mathbf{W}^{T}\mathbf{S}_{I}\mathbf{W})}
\end{equation}
with
\begin{small}
\begin{equation}\label{eq22}
\mathbf{S}_{O}=\sum_{j,l_j\neq p}(\mathbf{x}_{j}-\mathbf{m}_{p})(\mathbf{x}_{j}-\mathbf{m}_{p})^{T}, \:\:\:\:\:\:
\mathbf{S}_{I}=\sum_{j,l_j=p}(\mathbf{x}_{j}-\mathbf{m}_p)(\mathbf{x}_{j}-\mathbf{m}_p)^{T}
\end{equation}
\end{small}
denoting the out-of-class and in-class scatter matrices, respectively. The solution of (\ref{eq5}) is obtained by the eigenvectors corresponding to the $d_i$ largest eigenvalues of $\mathbf{S}_{I}^{-1}\mathbf{S}_{O}$.

The optimal dimensionality $d_i$ may vary for each class. For classes that are already highly discriminated from the others, fewer dimensions may be needed as compared to classes that are densely mixed with other classes. Since the rank of $\mathbf{S}_{I}$ is at most $n_p-1$ ($n_p$ is the number of samples from positive class), the dimensionality of the learnt subspace can be at most $\min(n_p-1,D)$.

%This resolves the flexibility limitation in the approach of LDA which utilizes a single shared subspace with dimensionality restricted to $C-1$.

\subsection{Multilinear Discriminant Analysis}
Several works have extended multi-class discriminant analysis criterion in order to utilize the natural tensor representation of the input data (\cite{kong2005two,ye2005two,yan2005discriminant,tao2007general,li2014multilinear}). We denote the set of $N$ tensor samples as $\{\mathcal{X}_1, \mathcal{X}_2, \dots, \mathcal{X}_N\}$, each with an associated class label $l_j$ ($j=1,\dots,N$) belonging to the label set $\{c_i \mid i=1,\dots,C\}$. The mean tensor of class $c_i$ is calculated as $\mathcal{M}_i=\frac{1}{n_i}\sum_{j=1}^{n_i}\mathcal{X}_{i,j}$ and the total mean tensor is $\mathcal{M}=\frac{1}{N}\sum_{i}^{C}\sum_{j=1}^{n_i}\mathcal{X}_{i,j}=\frac{1}{N}\sum_{i=1}^{C}n_i\mathcal{M}_i$. MDA seeks a set of projection matrices $\mathbf{W}_k\in \mathbb{R}^{I_k \times I_{k}^{'}}, I_{k}^{'}<I_k, k=1,\dots,K$ that map $\mathcal{X}_{i,j}$ to $\mathcal{Y}_{i,j}\in \mathbb{R}^{I_{1}^{'}\times \dots \times I_{K}^{'}}$, with the subspace projection defined as
\begin{equation}\label{eq6}
\mathcal{Y}_{i,j}=\mathcal{X}_{i,j}\prod_{k=1}^{K}\times_{k}\mathbf{W}_{k}^{T}
\end{equation}

Similar to LDA, the set of optimal projection matrices are obtained by maximizing ratio between the inter-class and intra-class distances, measured in the tensor subspace $\mathbb{R}^{I_{1}^{'}\times \dots \times I_{K}^{'}}$
\begin{equation}\label{eq7}
J(\mathbf{W}_{1},\dots,\mathbf{W}_K)=\frac{D_b}{D_w}
\end{equation}
where
\begin{equation}\label{eq8}
D_b=\sum_{i=1}^{C}\sum_{j=1}^{n_i}\big\Vert \mathcal{X}_{i,j}\prod_{k=1}^{K}\times_{k}\mathbf{W}_k^T-\mathcal{M}\prod_{k=1}^{K}\times_{k}\mathbf{W}_k^T\big\Vert_{F}^{2}
\end{equation}
\begin{equation}\label{eq9}
D_w=\sum_{i=1}^{C}\sum_{j=1}^{n_i}\big\Vert \mathcal{X}_{i,j}\prod_{k=1}^{K}\times_{k}\mathbf{W}_k^T-\mathcal{M}_{i}\prod_{k=1}^{K}\times_{k}\mathbf{W}_k^T \big\Vert_{F}^{2}
\end{equation}
are, respectively, the between-class and within-class distances.

An iterative approach is usually employed to solve the optimization problem in (\ref{eq7}). For example (\cite{li2014multilinear}) proposed CMDA algorithm that assumes orthogonal constraints on each projection matrix $\mathbf{W}_{k}^{T}\mathbf{W}_k=\mathbf{I}, k=1,\dots,K$ and optimizes (\ref{eq7}) by iteratively solving the following trace ratio problem for each mode-$k$
\begin{equation}\label{eq12}
J\big(\mathbf{W}_k\big)=\frac{tr\big(\mathbf{W}_{k}^{T} \mathbf{S}_{b}^{k} \mathbf{W}_{k}\big)}{tr\big(\mathbf{W}_{k}^{T} \mathbf{S}_{w}^{k} \mathbf{W}_{k}\big)}
\end{equation}
where
\begin{small}
\begin{equation}\label{eq10}
\mathbf{S}_{b}^{k}{} =\sum_{i=1}^{C} n_i \Bigg[(\mathcal{M}_i-\mathcal{M})\prod_{q\neq k}^{K}\times_{q}\mathbf{W}_{q}^{T}\Bigg]_{(k)}
 \Bigg[(\mathcal{M}_i-\mathcal{M})\prod_{q\neq k}^{K}\times_{q}\mathbf{W}_{q}^{T}\Bigg]_{(k)}^{T}
\end{equation}
\begin{equation}\label{eq11}
\mathbf{S}_{w}^{k}{} =\sum_{i=1}^{C} \sum_{j=1}^{n_i} \Bigg[(\mathcal{X}_{i,j}-\mathcal{M}_i)\prod_{q\neq k}^{K}\times_{q}\mathbf{W}_{q}^{T}\Bigg]_{(k)}
 \Bigg[(\mathcal{X}_{i,j}-\mathcal{M}_i)\prod_{q\neq k}^{K}\times_{q}\mathbf{W}_{q}^{T}\Bigg]_{(k)}^{T}
\end{equation}
\end{small}
are the between-class and within-class scatter matrices in mode-$k$.

CMDA first initializes $\mathbf{W}_k, k=1,\dots,K$ with all ones. At each iteration, the algorithm sequentially updates $\mathbf{W}_k$ by maximizing (\ref{eq12}) while keeping the rest projection matrices fixed.

\section{Multilinear Class-Specific Discriminant Analysis}\label{S:MCSDA}
In this section, we formulate the proposed multilinear version of CSDA, called Multilinear Class-Specific Discriminant Analysis (MCSDA). MCSDA finds a set of projection matrices that map $I_1\times \dots \times I_K$-dimensional tensor space to a smaller tensor as defined in (\ref{eq6}). The objective function of MCSDA is to find a tensor subspace in which the distances of the negative samples from the positive mean tensor are maximized and the distances of the positive samples from it are minimized.

Let us denote by $\mathcal{M}_p=\frac{1}{n_p}\sum_{j,l_j=p}\mathcal{X}_j$ the mean tensor of the positive class. The out-of-class and in-class distances are defined as follows
\begin{equation}\label{eq13}
\begin{split}
&D_O=\sum_{j,l_j\neq p}\big\Vert \mathcal{X}_{j} \prod_{k=1}^{K}\times_k \mathbf{W}_k^T-\mathcal{M}_p \prod_{k=1}^{K}\times_k \mathbf{W}_k^T \big\Vert_F^2 \\
&D_I=\sum_{j,l_j=p}\big\Vert \mathcal{X}_{j} \prod_{k=1}^{K}\times_k\mathbf{W}_k^T-\mathcal{M}_p \prod_{k=1}^{K}\times_k\mathbf{W}_k^T \big\Vert_F^2
\end{split}
\end{equation}
The MCSDA criterion is then expressed as
\begin{equation}\label{eq14}
J(\mathbf{W}_1,\dots,\mathbf{W}_K)=\frac{D_O}{D_I}.
\end{equation}

As in case of MDA, the objective in (\ref{eq14}) exposes a dependency between each $\mathbf{W}_k$. We therefore optimize (\ref{eq14}) by applying an iterative optimization process. In order to optimize for each $\mathbf{W}_k$, $D_O$ and $D_I$ need to be expressed as functions of $\mathbf{W}_k$. This can be done for $D_O$ by utilizing the relation in (\ref{eq1}), i.e.
\begin{small}
\begin{equation}\label{eq15}
\begin{split}
D_{O}^{k}{}=&\sum_{j,l_j \neq p}\big\Vert \mathbf{W}_{k}^{T}\big[\mathcal{X}_j \prod_{q\neq k}\times_q \mathbf{W}_q^T\big]_{(k)} -
\mathbf{W}_{k}^{T} \big[\mathcal{M}_p \prod_{q\neq k}\times_q \mathbf{W}_q^T\big]_{(k)} \big\Vert_{F}^{2}\\
%=&\sum_{c_i \neq p}\big\Vert \mathbf{W}_{k}^{T}\Big[\big(\mathcal{X}_{i}-\mathcal{M}_p\big)\prod_{q\neq k}\times_q\mathbf{W}_q^T\Big]_{(k)}\big\Vert_{F}^{2}\\
=&tr\bigg(\mathbf{W}_{k}^{T}\sum_{j,l_j\neq p}\Big[\big(\mathcal{X}_j-\mathcal{M}_p\big)\prod_{q\neq k}\times_q\mathbf{W}_q^T\Big]_{(k)}
 \Big[\big(\mathcal{X}_j-\mathcal{M}_p\big)\prod_{q\neq k}\times_q\mathbf{W}_q^T\Big]_{(k)}^{T}\mathbf{W}_k \bigg)
\end{split}
\end{equation}
\end{small}
where $D_{O}^{k}$ denotes $D_O$ after unfolding the projected tensor in mode-$k$. Let us denote $\mathbf{S}_{O}^{k}$ the out-of-class scatter matrix in mode-$k$, which is defined as
\begin{small}
\begin{equation}\label{eq17}
\mathbf{S}_{O}^{k}= \sum_{j,l_j\neq p}\Big[\big(\mathcal{X}_j-\mathcal{M}_p\big)\prod_{q\neq k}\times_q\mathbf{W}_q^{T}\Big]_{(k)}
\Big[\big(\mathcal{X}_j-\mathcal{M}_p\big)\prod_{q\neq k}\times_q\mathbf{W}_q^{T}\Big]_{(k)}^{T}
\end{equation}
\end{small}
Then $D_{O}^{k}$ in (\ref{eq15}) is expressed as $D_{O}^{k}=tr\big(\mathbf{W}_{k}^{T}\mathbf{S}_{O}^{k}\mathbf{W}_{k}\big)$. In a similar manner, the in-class distance calculated in mode-$k$ is expressed as $D_{I}^{k}=tr\big(\mathbf{W}_{k}^{T}\mathbf{S}_{I}^{k}\mathbf{W}_{k}\big)$ with
\begin{small}
\begin{equation}\label{eq18}
\mathbf{S}_{I}^{k}= \sum_{j,l_j= p}\Big[\big(\mathcal{X}_j-\mathcal{M}_p\big)\prod_{q\neq k}\times_q\mathbf{W}_q^T\Big]_{(k)}
\Big[\big(\mathcal{X}_j-\mathcal{M}_p\big)\prod_{q\neq k}\times_q\mathbf{W}_q^T\Big]_{(k)}^{T}
\end{equation}
\end{small}

Finally, the class-specific criterion with respect to $\mathbf{W}_k$
\begin{equation}\label{eq19}
J(\mathbf{W}_k)=\frac{tr\big(\mathbf{W}_{k}^{T}\mathbf{S}_{O}^{k}\mathbf{W}_k\big)}{tr\big(\mathbf{W}_{k}^{T}\mathbf{S}_{I}^{k}\mathbf{W}_k\big)}
\end{equation}
MCSDA starts by initializing $\mathbf{W}_k$ with ones. At each iteration, it updates $\mathbf{W}_{k}$ by maximizing (\ref{eq19}), while keeping the rest projection matrices fixed. A detailed description of the MCSDA optimization process is presented in Algorithm 1.

\subsection{Complexity Discussion}
It is clear that the number of parameters of the tensor version is much lower compared to the vector version. Suppose the dimensionality of each tensor sample is $\mathbb{R}^{I_1 \times \dots \times I_K}$, which corresponds to a vectorized sample in $\mathbb{R}^{\prod_{k=1}^{K}I_k}$. Given the tensor subspace is $\mathbb{R}^{I_1^{'} \times \dots \times I_K^{'}}$, the number of parameters for MCSDA is equal to $\sum_{k=1}^{K}I_{k} I_{k}^{'}$. The corresponding CSDA model projects each vectorized sample from $\mathbb{R}^{\prod_{k=1}^{K} I_{k}}$ to $\mathbb{R}^{\prod_{k=1}^{K}I_{k}^{'}}$, requiring $\prod_{k=1}^{K}I_{k}I_{k}^{'}$ parameters. In order better understand the difference between the two cases, let us consider the following example. For an image of size $30\times30$ pixels projected to a scalar, the tensor model learns $30\times1+30\times1=60$ parameters, while the vector model learns $30\times30=900$ parameters.

Regarding time complexity, let us denote $I=\prod_{k=1}^{K}I_k$ the total number of elements in input space and $I^{'}=\prod_{k=1}^{K}I_{k}^{'}$ the total number of elements in the learnt subspace. The solution of CSDA involves the following steps:
\begin{itemize}
	\item Calculation of $\mathbf{S}_O$ and $\mathbf{S}_I$ defined in (\ref{eq22}) having time complexity of $O(NI^2)$.
	\item Calculation of $\mathbf{S}_{I}^{-1}\mathbf{S}_O$ requires matrix inversion of $\mathbf{S}_I$ and matrix multiplication between $\mathbf{S}_{I}^{-1}$ and $\mathbf{S}_O$, having time complexity of $O(2I^3)$.
	\item Eigenvalue decomposition of $\mathbf{S}_{I}^{-1}\mathbf{S}_{O}$ having time complexity of $O(\frac{9}{2}I^3)$.
\end{itemize}
Thus the total time complexity of CSDA is
\begin{equation}\label{eq20}
O(NI^2+\frac{13}{2}I^3)
\end{equation}

\begin{algorithm}[t!]
	\caption{MCSDA}
	\textbf{Input}: Training tensor $\mathcal{X}_{i,j}\in \mathbb{R}^{I_1\times \dots \times I_K}$ and respective labels $c_i$; Subspace dimensionality $I_{1}^{'}\times \dots \times I_{k}^{'}$; maximum iteration $\tau$ and threshold $\epsilon$.
	\begin{algorithmic}[1]
		\State Initialize $\mathbf{W}_k(0), k=1,\dots,K$ with $\mathbf{1}$
		\For{$t \leftarrow 1$ to $\tau$}
		\For{$k \leftarrow 1$ to $K$}
		\State Calculate $\mathbf{S}_{O}^{k}$ according to (\ref{eq17}) using $\mathbf{W}_{k}(t-1)$
		\State Calculate $\mathbf{S}_{I}^{k}$ according to (\ref{eq18}) using $\mathbf{W}_{k}(t-1)$
		\State Update $\mathbf{W}_{k}(t)$ by solving (\ref{eq19})
		\EndFor
		\State \textbf{end for}
		\If { $\sum_{k=1}^{K}\big\Vert \mathbf{W}_{k}(t)\mathbf{W}_{k}^{T}(t-1)-\mathbf{I}\big\Vert_{F} \leq \epsilon$}
		\State Terminate
		\EndIf
		\State \textbf{end if}
		\EndFor
		\State \textbf{end for}
	\end{algorithmic}
	\textbf{Output}: Projection matrices $\mathbf{W}_{k}, \:k=1,\dots, K$
\end{algorithm}

MCSDA employs an iterative process parameterized by the terminating threshold $\epsilon$ and the number of maximum iteration $\tau$. At each iteration, MCSDA requires the following computation steps:
\begin{itemize}
	\item Calculation of $\mathbf{S}_{I}^k$ and $\mathbf{S}_{O}^k$ requires the projection of $\mathcal{X}_j$ to $\mathbb{R}^{I_1^{'}\times \dots \times I_k \times \dots \times I_K^{'}}$ having time complexity of $O(NI_k^{'}I)$.
	\item Calculation of $(\mathbf{S}_{I}^{k})^{-1}\mathbf{S}_{O}^k$ requires matrix inversion of $\mathbf{S}_{I}^{k}$ and matrix multiplication between $(\mathbf{S}_{I}^{k})^{-1}$ and $\mathbf{S}_{O}^{k}$, having time complexity of $O(2I_k^3)$.
	\item Eigenvalue decomposition of $(\mathbf{S}_{I}^{k})^{-1}\mathbf{S}_{O}^k$ having time complexity of $O(\frac{9}{2}I_k^3)$.
\end{itemize}
Hence, the computational cost to update $\mathbf{W}_k$ of MCSDA is $O(NI_k^{'}I+\frac{13}{2}I_k^3)$. Let $\tau$ be the number of maximum iteration, the maximum cost of computation of MCSDA is
\begin{equation}\label{eq21}
O(\tau N I\sum_{k=1}^{K}I_k^{'}+\frac{13}{2}\tau\sum_{k=1}^{K}I_k^3)
\end{equation}

Due to the iterative nature of MCSDA, it is not straightforward to compare the time complexity of MCSDA with that of CSDA. Our experiments showed that with the maximum number of iteration set to $\tau=20$, MCSDA already achieves good performance. In addition, for frequently encountered data, the number of tensor modes $K$ ranging from $2$ to $4$. For example, grayscale images, EEG multichannel data or time-series financial data has $K=2$ while RGB images has $K=3$ or video data has $K=4$. Comparing the first two terms of (\ref{eq20}) and (\ref{eq21}) and noting the fact that the dimensions of the projected space are usually much smaller than the input, it is easy to see that $NI^2=NI\prod_{k=1}^{K}I_k > NI\tau\sum_{k=1}^{K}I_k^{'}$. Comparing the second term of (\ref{eq20}) and (\ref{eq21}), it is also clear that $\frac{13}{2}I^3=\frac{13}{2}\prod_{k=1}^{K}I_k^3>\frac{13}{2}\tau\sum_{k=1}^{K}I_k^3$.

To conclude, the solution of the vector model is more costly in terms of computation as compared to the tensor model. Moreover, the vector approach with $O(I^3)$ becomes impractical when $I$ scales to the order of thousands or more, which is usually the case. In contrast, the tensor approach with $O(I_k^3)$ is scalable with high-dimensional input.

\section{Experiments}\label{S:Experiments}
In this section, we provide experiments conducted in order to evaluate the effectiveness of the proposed MCSDA and compare it with related discriminant analysis methods, namely vector-based Class-Specific Discriminant Analysis (CSDA) and Mulitilinear Discriminant Analysis (MDA) (\cite{li2014multilinear}). It should be noted that the class-specific methods model $C$ classes as $C$ binary problems, we therefore conducted the experiments in which $C$ one-vs-rest MDA classifiers are learned. We performed the benchmark in three publicly available datasets coming from two application domains: face verification and stock price prediction based on limit order book data. Detailed description of the datasets and the corresponding experimental settings are provided in the following subsections.

Since all the competing methods are subspace methods, after learning the optimal projection matrices, one can train any classifier on the data representations in the discriminant subspace to boost the performance. For example, the distance between training sample and each class mean vectors can be used as the training data for SVM classifier. In the test phase, a test sample is projected to the discriminant subspace and distances between test sample and each class mean are used as feature vector fed to the learnt SVM classifier, similar to (\cite{iosifidis2015class}). Since the goal of this paper is to directly compare the discrimination ability of MCSDA, compared to that of CSDA and MDA, we do not train any other classifier in the discriminant space. In the test phase, the similarity score is calculated as the inverse of the distance between the test sample and the positive mean in the discriminant space. The similarity scores are used to evaluate the performance of each algorithm, based on different metrics as described next.

\subsection{Facial Image Datasets}
Since tensor is a natural representation of image data, we employ two facial image datasets, namely ORL and CMU PIE, with different sizes to compare the performance of the tensor-based and vector-based methods.
The ORL dataset (\cite{samaria1994parameterisation}) consists of $400$ facial images depicting $40$ persons ($10$ images each). The images were taken at different times with different conditions in terms of lighting, facial expressions (smile/neutral) and facial details (open/closed eyes, with/without glasses). All of the images were captured in frontal position with a tolerance of rotation and tilting up to $20$ degrees. The CMU PIE dataset (\cite{sim2002cmu}) consists of $64$ individuals with $41,368$ facial images in total. The images were captured with $13$ different camera positions and $21$ flashes under different pose, illumination and expression. All images in 5 near frontal positions ($C05, C07, C09, C27, C29$) of 8 individuals ($55, 57, 58, 59, 61, 66, 67, 68, 69$) were used in our experiments. Moreover, all images used in our experiments are in grayscale format.

Using the above datasets, we formulate multiple face verification problems. That is, a class-specific model is learned for a person of interest, either using class-specific or the multi-class (in this case binary) criterion. During the test phase, image a test image is presented and the model decides whether the image depicts the person of interest or not (\cite{goudelis2007class,iosifidis2016class,iosifidis2016tifs}). We measure the performance of each model by calculating the Average Precision (AP) metric. This process is applied multiple times (equal to the number of persons in each dataset) and the performance of each approach is obtained by calculating the mean Average Precision (mAP) over all sub-experiments. We applied multiple experiments based on five different train/test split sizes, where $k$ percent of the data is randomly selected for training and the rest for testing with $k\in \{0.1, 0.2, 0.25, 0.35, 0.5\}$. For each value of $k$, 5 experiments were repeated and the average result is reported.

Regarding the preprocessing step, all facial images were cropped and resized to $40\times 30$ pixels. For tensor-based approaches, we keep the projected dimension of both mode-$1$ and mode-$2$ equal, ranging from $2$ to $20$. The maximum number of iterations is set to $\tau=20$ and the terminating threshold $\epsilon=1e-5$. To ensure stability, we regularized $\mathbf{S}_w^k$ in MDA, $\mathbf{S}_{I}^k$ in MCSDA and $\mathbf{S}_I$ in CSDA by adding a scaled version of the identity matrix (using a value of $\lambda=0.01$). We also investigated the case when additional information is available by generating HOG images (\cite{dalal2005histograms}) from the original images and concatenating the original image and its HOG image to form a $3$-mode tensors of size $40\times 30 \times 2$. The results from the enriched version are denoted by CSDA-H, MCSDA-H and MDA-H.

\subsection{Limit Order Book Dataset}
In addition to image data, (multi-dimensional) time-series, like limit order book (LOB) data, also have a natural representation as tensors of two modes. In our experiments, a recently introduced LOB dataset, called FI-2010 (\cite{ntakaris2017benchmark}), was used. FI-2010 collected order information of 5 Finnish stocks in 10 consecutive days, resulting in more than 4 millions of order events. For every $10$ consecutive order events a $144$-dimensional feature vector is extracted and a corresponding label is defined indicating the prospective change (increase, decrease or stationary) of the mid-price after the next $10$ order events. For the vector models, each sample is of size $144$ dimension, representing information from $10$ most recent order events. In order to take into account more information in the recent past, our tensor models exploits a tensor sample of size $144\times 10$, representing information from $100$ most recent order events.

We followed the standard day-based anchored cross-validation sets provided by the database with 9 folds in total. For the tensor-based models, we varied the projected dimension of the first mode from $5$ to $60$ with a step of $5$ and the second mode from $1$ to $8$. The values of $\tau$ and $\epsilon$ were the same as those used in the face verification experiments. Since FI-2010 is an unbalanced dataset with the mid-price remaining stationary most of the time, we cross-validated based on average $f1$ score per class and also report the corresponding accuracy, average precision per class, average recall per class. Since our experimental protocol is the same with that used in (\cite{passali2017time}) for the Bag-of-Words (BoF) and Neural Bag-of-Words (N-BoF) models, we directly report their results. In addition, we report the baseline results from the database (\cite{ntakaris2017benchmark}) using Single Layer Feed-forward Network (SLFN) and Ridge Regression (RR).

\subsection{Results}\label{Results}
The results from 2 facial datasets are presented in Table \ref{t1} and Table \ref{t2}. Moreover, the last column of Tables 1 and 2 shows the relative computation time ($t$) of each method measured on the same machine (normalized with respect to the computation time of the proposed MCSDA method). Comparing the vector model and the tensor model utilizing class-specific criterion, it is clear that CSDA slightly outperforms the proposed MCSDA. However, as can be seen, the computational time (normalized with respect to the training time of MCSDA) of CSDA is higher (by one or two orders of ten). The computational efficiency of the proposed MCSDA over CSDA becomes more significant when the dimension of the input scales up. While the number of elements in the input doubles, computation time of MCSDA-H scales favourably while CSDA-H requires approximately $7$ times more computation compared to CSDA. The result justifies our analysis in the time complexity discussion section above. Comparing the two tensor-based approaches, the proposed MCSDA outperforms MDA in most of the configurations of $k$, while their computational times are similar. Regarding the exploitation of enriched information, we can observe that all competing methods achieved some improvements. The benefit of additional information is marginal when the training data is small but clearly visible when $50$\% of the data is used for training for the tensor-based methods. In contrast, the benefit of additional information for the vector-based model is very small.

The results for stock prediction using LOB data are presented in Table \ref{t3}. While the performance of MCSDA was not better than its vector counterpart in the above face verification experiments, MCSDA outperforms all competing methods in the stock prediction problem, including the more complex neural network-based bag-of-words model N-BoF (\cite{passali2017time}).

The difference in the relative performance between the vector-based and tensor-based variants in the two different application domains can be explained by looking into the optimal dimensionality of the subspaces obtained for both CSDA and MCSDA. In the two image verification problems, the optimal dimensionality of the subspace obtained for MCSDA is equal to $7\times7=49$ dimensions for both ORL and CMU PIE datasets. For CSDA, the optimal subspace dimensionality is equal to $27\times27=729$ dimensions for ORL and $25\times25=625$ dimensions for CMU PIE. This result shows that in the CSDA case, the number of parameters is much higher compared to its tensor counterpart. In facial images, several visual cues are usually necessary to perform the verification. Since the vector approach estimates many more parameters, more visual cues can be captured, which leads to better performance, compared to MCSDA. However, this comes with a much higher computational cost.

In the stock prediction problem, the difference between the number of estimated parameters for MCSDA and CSDA is small. Particularly, over $10$ folds the average number of parameters estimated for MCSDA is approximately equal to $6300$, while for CSDA is slightly over and equal to $6000$. Since multilinear class-specific projection (MCSDA) can perform the projection along temporal mode (mode-$2$) too, MCSDA can potentially capture important temporal cues required to predict future movements in stock price. The experiment in stock price prediction problem shows the potential of multilinear techniques in general, and MCSDA in particular, in exploiting the multilinear structure of the time-series data.

\begin{table}[]
	\begin{center}
		\caption{Performance (mAP) on ORL dataset}
		\resizebox{\linewidth}{!}{
			\begin{tabular}{|l|c|c|c|c|c|c|}\cline{2-7}
				\multicolumn{1}{c|}{}
				& $k=0.1$ 	& $k=0.2$ 	& $k=0.25$ 	& $k=0.35$ 	& $k=0.5$  &$t$\\ \hline
				CSDA	    & $76.81$	& $87.08$	& $91.42$	& $93.72$	& $97.81$ 	&$\mathit{12}$\\ \hline		
				MDA 	    & $63.99$	& $75.21$	& $79.26$	& $81.41$	& $88.41$ 	&$\mathit{1.17}$\\ \hline
                \bf MCSDA 	& $72.20$ 	& $84.73$ 	& $87.65$	& $92.05$	& $95.69$ 	&$\mathit{1}$\\ \hline \hline
				CSDA-H	    & $77.00$	& $87.21$	& $91.50$	& $93.80$	& $97.87$ 	&$\mathit{92.10}$\\ \hline		
				MDA-H	    & $64.69$	& $75.84$	& $80.18$	& $82.80$	& $89.52$ 	&$\mathit{1.83}$\\ \hline
                \bf MCSDA-H	& $72.21$ 	& $84.54$ 	& $87.97$	& $92.36$	& $96.27$ 	&$\mathit{1.4}$\\ \hline
			\end{tabular}\label{t1}
		}
	\end{center}
\end{table}
\begin{table}[]
	\begin{center}
		\caption{Performance (mAP) on CMU PIE dataset}
		\resizebox{\linewidth}{!}{
			\begin{tabular}{|l|c|c|c|c|c|c|}\cline{2-7}
				\multicolumn{1}{c|}{}& $k=0.1$ 	& $k=0.2$ 	& $k=0.25$ 	& $k=0.35$ 	& $k=0.5$	&$t$ \\ \hline
				CSDA	    & $76.99$	& $89.49$	& $93.09$	& $93.91$	& $95.76$ 	&$\mathit{12.56}$\\ \hline		
				MDA 	    & $79.77$	& $88.29$	& $89.67$	& $90.07$	& $91.20$ 	&$\mathit{1.02}$\\ \hline
                \bf MCSDA 	& $79.88$ 	& $89.06$ 	& $90.36$	& $91.45$	& $92.57$ 	&$\mathit{1}$\\ \hline \hline
				CSDA-H	    & $76.99$	& $90.17$	& $93.08$	& $94.46$	& $95.95$ 	&$\mathit{72.86}$\\ \hline		
				MDA-H       & $82.26$	& $88.69$	& $90.89$	& $91.28$	& $92.66$ 	&$\mathit{2.91}$\\ \hline
                \bf MCSDA-H	& $80.71$ 	& $89.41$ 	& $90.07$	& $92.20$	& $93.88$ 	&$\mathit{2.81}$\\ \hline  				
			\end{tabular}\label{t2}
		}
	\end{center}
\end{table}
\begin{table}[]
	\begin{center}
		\caption{Performance on FI-2010 dataset}
		\resizebox{\linewidth}{!}{
			\begin{tabular}{|l|c|c|c|c|}\cline{2-5}
				\multicolumn{1}{c|}{}
				& Accuracy 			& Precision			& Recall			& F1 		\\ \hline
				RR		    & $46.00\pm 2.85$	& $43.30\pm 9.9$	& $43.54\pm 5.2$	& $42.52\pm 1.22$	\\ \hline		
				SLFN	    & $53.22\pm 7.04$	& $49.60\pm 3.81$	& $41.28\pm 4.04$	& $38.24\pm 5.66$	\\ \hline \hline
				CSDA	    & $79.15\pm 4.89$	& $40.20\pm 0.65$	&$42.91\pm 1.87$	& $40.79\pm 0.69$	\\ \hline
				MDA		    & $83.92\pm 3.23$	& $45.08\pm 1.37$	&$45.92\pm 1.64$	& $45.33\pm 1.33$	\\ \hline	
                \bf MCSDA	& $83.66\pm 3.69$	& $46.11\pm 1.26$	&$48.00\pm 1.63$	& $\mathbf{46.72}\pm 1.29$	\\ \hline \hline
				BoF		    & $57.59\pm 7.34$	& $39.26\pm 0.94$	&$51.44\pm 2.53$	& $36.28\pm 2.85$	\\ \hline
				N-BoF	    & $62.70\pm 6.73$	& $42.28\pm 0.87$	&$61.41\pm 3.68$	& $41.63\pm 1.90$	\\ \hline
			\end{tabular}\label{t3}
		}
	\end{center}
\end{table}

\section{Conclusions}\label{S:Conclusions}
In this paper, we proposed a tensor subspace learning method that utilizes the intrinsic tensor representation of the data together with the class-specific discrimination criterion. We provided a theoretical discussion of the time complexity of the proposed method, compared with its vector counterpart. Experimental results show that the MCSDA is computationally efficient and scalable with performance close to its vector counterpart in face verification problems, while outperformed other competing methods in a stock price prediction problem problem based on Limit Order Book data.

\bibliography{cstda}
\bibliographystyle{ieeetr}

\end{document}